\documentclass[letterpaper, 10 pt, conference]{ieeeconf}  %

\IEEEoverridecommandlockouts                              

\usepackage{graphics} 
\usepackage{epsfig} 
\usepackage{mathptmx} 
\usepackage{times} 
\usepackage{amsmath} 
\usepackage{amssymb}  
\usepackage{multirow, multicol, booktabs, microtype}
\usepackage{subcaption}
\usepackage{hyperref}
\usepackage{authblk}
\usepackage{balance}

\title{\LARGE \bf
Class-N-Diff: Classification-Induced Diffusion Model Can Make Fair Skin Cancer Diagnosis}

\author{Nusrat Munia} 
\author{Abdullah Imran}

\affil{Department of Computer Science\\
University of Kentucky, Lexington, KY 40506, USA
}

\begin{document}

\maketitle
\thispagestyle{empty}
\pagestyle{empty}

\begin{abstract}

Generative models, especially Diffusion Models, have demonstrated remarkable capability in generating high-quality synthetic data, including medical images. However, traditional class-conditioned generative models often struggle to generate images that accurately represent specific medical categories, limiting their usefulness for applications such as skin cancer diagnosis. To address this problem, we propose a classification-induced diffusion model, namely, Class-N-Diff, to simultaneously generate and classify dermoscopic images. Our Class-N-Diff model integrates a classifier within a diffusion model to guide image generation based on its class conditions. Thus, the model has better control over class-conditioned image synthesis, resulting in more realistic and diverse images. Additionally, the classifier demonstrates improved performance, highlighting its effectiveness for downstream diagnostic tasks. This unique integration in our Class-N-Diff makes it a robust tool for enhancing the quality and utility of diffusion model-based synthetic dermoscopic image generation. Our code is available at \url{https://github.com/Munia03/Class-N-Diff}.
\newline


\end{abstract}

\section{INTRODUCTION}

Accurate skin cancer diagnosis is one of the biggest challenges in medicine. Artificial intelligence (AI), more specifically, deep learning models, have shown remarkable effectiveness in the early detection and diagnosis of skin cancer~\cite{imran2024enhanced,wu2022skin, chaturvedi2020multi, hosny2018skin}. These models can achieve high accuracy in detecting malignant and benign skin lesions when trained on large dermoscopic image datasets. However, their performance can be biased, as they fail to generalize across different subgroups, e.g., skin tones~\cite{tschandl2018ham10000, kinyanjui2020fairness, groh2021fitzpatrick}. Such disparities arise due to imbalanced datasets where lighter skin tones are often overrepresented, leading to poorer performance for underrepresented groups. This can lead to significant healthcare inequities and affect underrepresented populations. 

Several works exist in the literature to address the fairness issue in disease diagnosis~\cite{ghadiri2024xtranprune, li2024iterative,bevan2022detecting,wu2022fairprune, daneshjou2022disparities, chiu2023toward, xu2023fairadabn}. For example, an ensemble mechanism combines separate models trained for lighter and darker skin tones~\cite{almuzaini2023toward}. A de-biasing technique has been proposed to remove skin tone features from dermatology images to reduce skin tone bias~\cite{bevan2022detecting}. Another work, FairAdaBN~\cite{xu2023fairadabn}, utilizes adaptive batch normalization for sensitive attributes, incorporating a loss function that reduces the disparity in prediction probabilities across subgroups. Different pruning techniques have been proposed where sensitive nodes are pruned from the model to eliminate dependence on sensitive attributes to mitigate biases~\cite{ghadiri2024xtranprune, wu2022fairprune, chiu2023toward}. FairSkin framework~\cite{zhang2024fairskin} uses a three-level resampling mechanism to ensure fairer representation across racial and disease categories. Although these methods have shown promising results, their effectiveness is limited by the lack of sufficient data from underrepresented populations.

Recent advancements in generative AI, particularly conditional diffusion models, have demonstrated promising results in medical image synthesis and analysis \cite{munia2025prompting, ktena2024generative, imran2021multi, munia2025dermdiff}. These models offer a new paradigm for mitigating bias in skin disease classification by generating diverse and balanced datasets. In this paper, we propose a novel approach that integrates a class-conditional diffusion model with a classifier to enhance fairness in deep learning-based skin lesion classification. By leveraging both generative and classification models, we create a more equitable and robust system for skin cancer detection across diverse populations. Our main contributions include:
\begin{itemize}
    \item A generative model framework (Class-N-Diff) conditioned on class labels with the integration of a classification model.
  
    \item A skin disease diagnosis model trained during the diffusion process to perform fairly across different subgroups.
    
    \item Our extensive experimental evaluation demonstrates improvements in both classification and generative performance by Class-N-Diff.
   
\end{itemize}

\begin{figure}[t]
    \centering
    \includegraphics[width=\linewidth, trim={2.5cm 0cm 2.5cm 0cm}, clip]{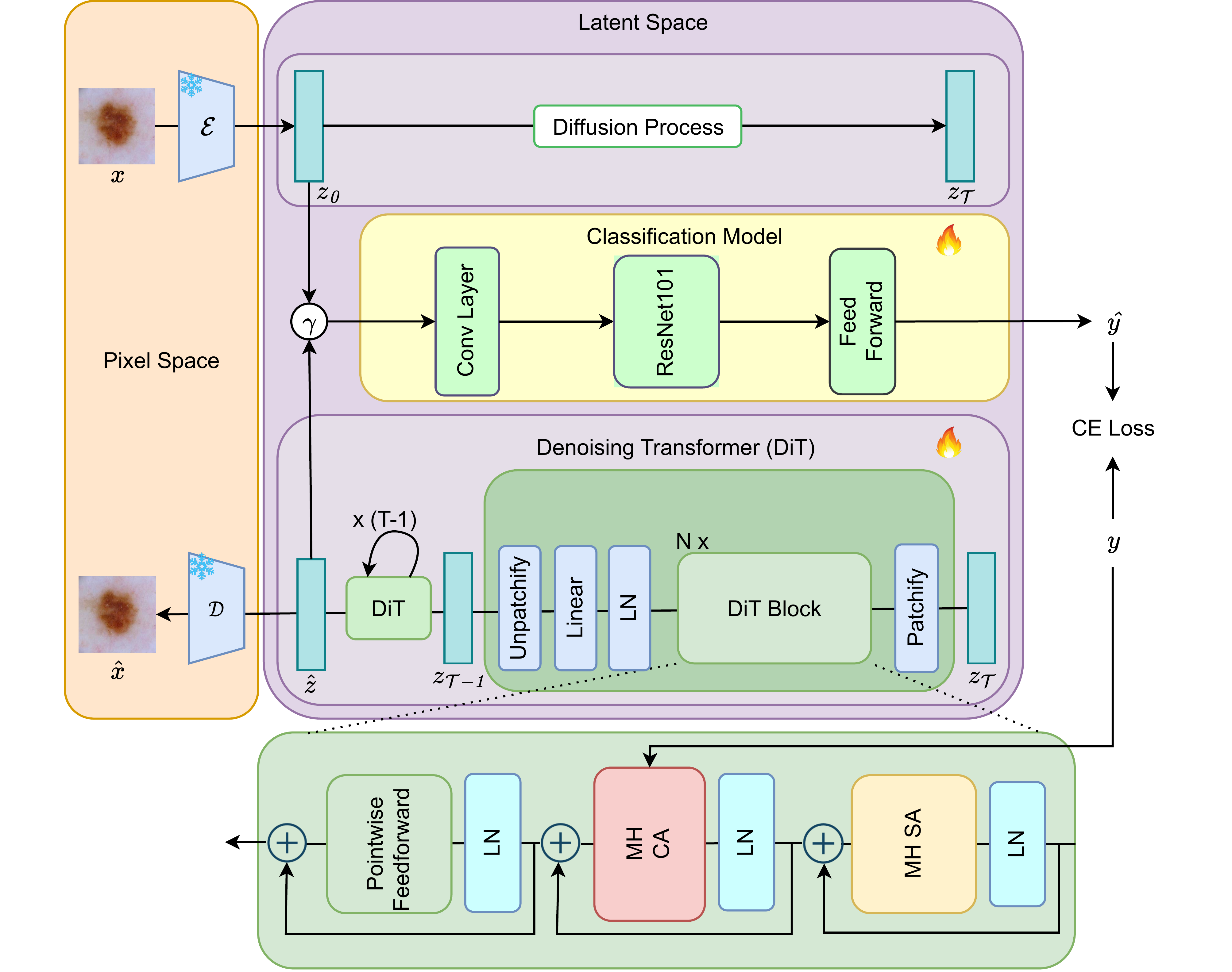}
    \caption{Proposed Class-N-Diff: Classification model induced class-conditioned transformer-based diffusion model that jointly performs classification and image generation. The encoder maps an input image to a latent space where the diffusion process adds noise, and the Denoising Transformer (DiT) reconstructs clean representations guided by class-aware attention. The ResNet-101 classifier provides class-conditioning and supervision via cross-entropy loss, enabling both accurate prediction and realistic dermoscopic image synthesis.}
    \label{fig:arch1}
\end{figure}

\section{Related Works}
In recent years, generative models such as Generative Adversarial Networks (GANs)~\cite{goodfellow2014generative, imran2021multi} and Diffusion Models~\cite{rombach2022high, ho2020denoising} have gained significant attention for their ability to generate realistic images. Generative models have been utilized in mitigating the biases in skin cancer diagnosis models. GAN-based augmentation has been used to reduce common artifact biases~\cite{mikolajczyk2022biasing}, such as hair, rulers, and image frames, but it overlooks the deeper sources of bias related to race and demographic diversity. Alternatively, diffusion models, such as the U-Net-based Stable Diffusion model~\cite{rombach2022high} and the transformer-based Diffusion Transformer~\cite{peebles2023scalable}, have achieved notable performance improvements in generating high-quality images that outperform GAN-based models.

A diffusion-based generative model has been employed to synthesize samples from underrepresented groups during the training of the disease classification model to mitigate the bias~\cite{li2024iterative}. This work trained an unconditional diffusion model and generated random samples from it. Diffusion models can, however, be conditioned on class labels, text prompts, or others for more control over the image generation process. For instance, DermDiff~\cite{munia2025dermdiff} utilizes a U-Net-based dermatology diffusion model conditioned on generic text prompts to generate dermoscopic images focusing on underrepresented groups. The text prompts contain both diagnosis and demographic information to provide the diffusion model with more detailed context on image generation. We propose to further improve the image generation in a diffusion model by incorporating a classification model. This joint learning approach can create a synergistic effect, enhancing both image generation quality and classification performance~\cite{imran2019multi}.

\section{Methods}

Fig.~\ref{fig:arch1} provides a visual illustration of our proposed Class-N-Diff. The framework is composed of two core components: a classification model and a diffusion model. 

\subsection{Classification model}
We assume to have a skin disease image $x$ and its class label $y$, where $y=0$ for the benign class and $y=1$ for the malignant class. To avoid large memory and computational bottlenecks, our proposed Class-N-Diff model operates in a compressed latent space~\cite{rombach2022high} rather than the high-dimensional pixel space for both the diffusion process and classification tasks. The input dermoscopic image $x$ is first encoded into a latent representation $z_0$ using a pre-trained variational autoencoder (VAE) model \cite{kingma2013auto}.
\begin{equation}
    z = VAE(x).
\end{equation}
This transformation reduces the dimensionality of the input image while keeping its essential features.
Then the latent representation $z$ goes through our classification model depending on the hyper-parameter $\gamma$ to predict the class label. The classification model starts with a convolutional layer to capture local spatial features, followed by a ResNet101~\cite{he2016deep} backbone to learn hierarchical representations for classification. A feed-forward layer then processes the features extracted by the ResNet101 model, and a sigmoid activation function is applied to get the final class prediction $\hat{y}$ for the input image $x$. We use a cross-entropy loss to optimize the classification model.
\begin{equation}
    \mathcal{L}_{CL} = CE (y, \hat{y}).
\end{equation}

\subsection{Diffusion Model}
We adopt the Diffusion Transformer (DiT)~\cite{peebles2023scalable} model in Class-N-Diff, which is built upon the Denoising Diffusion Probabilistic Models (DDPM) framework~\cite{ho2020denoising}. The diffusion model consists of two stages: a forward (diffusion) process that gradually corrupts the input image with noise and a reverse (denoising) process that iteratively removes that noise to reconstruct the original image.

\noindent\textbf{Forward process (Diffusion):}  For a dermoscopic image $x$, the forward process progressively adds noise to the image latent $z$ with $t$ timestep: $q(z_t|z) = \mathcal{N}(z_t; \sqrt{\bar{\alpha}_t}z, (1 - \bar{\alpha}_t)\mathbf{I})$, where $\bar{\alpha}_t$ are hyperparameters. With reparameterization, we sample $z_t = \sqrt{\bar{\alpha}_t}z + \sqrt{1 - \bar{\alpha}_t}\epsilon_t$, where $\epsilon_t \sim \mathcal{N}(0, \mathbf{I})$.

\begin{table*}[t]
\centering
\caption{Different experimental Settings for evaluating the Class-N-Diff model performance.}
\begin{tabular}{@{}lcp{12cm}@{}}
\toprule
\textbf{Model} & \textbf{Setting} & \textbf{Description} \\
\midrule
Class Conditional DiT & Setting 1 & Class-conditioned diffusion model without the classification model. \\
\midrule
\multirow{4}{*}{Class-N-Diff}  & Setting 2 & Train classification model for the last two epochs only with $\gamma = 0.25$, and $\lambda = 0.2$. \\

& Setting 3 & Periodically increase the value of $\gamma$ starting from 0 and $\lambda = 0.2$. Optimizer step: once in three steps. \\

& Setting 4 & Periodically increase the value of $\gamma$ starting from 0 and $\lambda = 0.2$. Optimizer step: each step. \\

& Setting 5 & Periodically increase the value of $\gamma$ starting from 0 and $\lambda = 0.3$. Optimizer step: once in three steps. \\
\bottomrule
\end{tabular}
\label{tab:experimental_settings}
\end{table*}

\noindent\textbf{Reverse process (Denoising):} The diffusion transformer learns the reverse process conditioned on class label $y$ to predict the noiseless latent from the noisy latent, i.e., $p_\theta(z_{t-1} | z_t, y)$. The transformer-based architecture learns to predict the mean $\mu_\theta(z_t)$ and the variance $\Sigma_\theta(z_t)$ of this reverse process. To train this reverse process, the mean $\mu_\theta$ is reparameterized to predict a noise $\epsilon_\theta$ by the model. The model is trained by minimizing the mean-squared error (MSE) between the predicted noise $\epsilon_\theta(z_t)$ and the true noise $\epsilon_t$, which is sampled from a standard Gaussian distribution.
\begin{equation}
   \mathcal{L}_{\text{diff}}(\theta) = \|\epsilon_\theta(z_t) - \epsilon_t\|_2^2. 
\end{equation}
We incorporate the DiT blocks with a multi-head cross-attention mechanism. Following~\cite{peebles2023scalable}, Class-N-Diff applies a patchification process to the noisy image $z_t$ in the latent space, which is then processed through multiple DiT blocks. The noise timestep $t$ and label embedding $y$ are concatenated and fed into the multi-head cross-attention layer within the DiT block for conditioning. The DiT block then takes the noise input $z_t$ and applies multi-head self-attention on $z_t$, as illustrated in Fig.~\ref{fig:arch1}.

A representation $\hat{z}$ of the de-noised latent image is obtained from the diffusion transformer model and passed to the classification model, depending on a gating variable $\gamma$. After computing a classification loss, it is added to the diffusion loss $\mathcal{L}_{\text{diff}}(\theta)$ using a weight parameter $\lambda$. The final loss for our diffusion process is, therefore, calculated as:
\begin{equation}
   \mathcal{L} = \mathcal{L}_{\text{diff}}(\theta) + \lambda * \mathcal{L}_{CL}. 
\end{equation}
The gating parameter $\gamma$ controls the input to the classification model; either the latent from the original image or the reconstructed image is passed to the model. Considering the vulnerability of the model initially during training, we set $\gamma$ to 0; that means only the original image latent is used for calculating the classification loss. Over the training period, the value of $\gamma$ is increased periodically to incorporate the latent from the diffusion model. Once the model is trained, it can generate new data by initializing from random noise $z_{\text{max}} \sim \mathcal{N}(0, I)$ and iteratively transforming this noise using the learned reverse denoising process. A learned decoder~\cite{kingma2013auto} is used to generate the new dermoscopic image from the newly generated latent: $ \hat{x} = D(\hat{z}) $. In Class-N-Diff, the DiT block is repeated 24 times, using a patch size of 4.

\begin{table*}[t]
\centering
\caption{Class-N-Diff Generative model evaluation: FID ($\downarrow$) and MS-SSIM Scores ($\downarrow$).}
\label{tab:fid_scores}
\resizebox{0.7\linewidth}{!}{%
\begin{tabular}{@{}llc c c c c@{}}
\toprule
\textbf{Model} & \textbf{Setting} & \textbf{FID (5k)} & \textbf{FID (10k)} & \textbf{FID (20k)} & \textbf{MS-SSIM} \\ 
\midrule
Class Conditional DiT & Setting 1 & 69.100 & 48.750 & 45.770 & 0.583\\ 
\midrule
\multirow{4}{*}{Class-N-Diff}   & Setting 2 & 27.210 & 15.940 & 18.270 & 0.372\\ 
    & Setting 3 & 3.930 & 2.710 & \textbf{2.420} & 0.316\\ 
    & Setting 4 & \textbf{2.690} & \textbf{2.640} & 2.750 & 0.462\\ 
    & Setting 5 & 4.290 & 3.900 & 2.430 & \textbf{0.285}\\ 
\bottomrule
\end{tabular}
}
\end{table*}

\begin{figure*}[t]
  \centering
    \subcaptionbox{}{\includegraphics[width=0.49\textwidth]{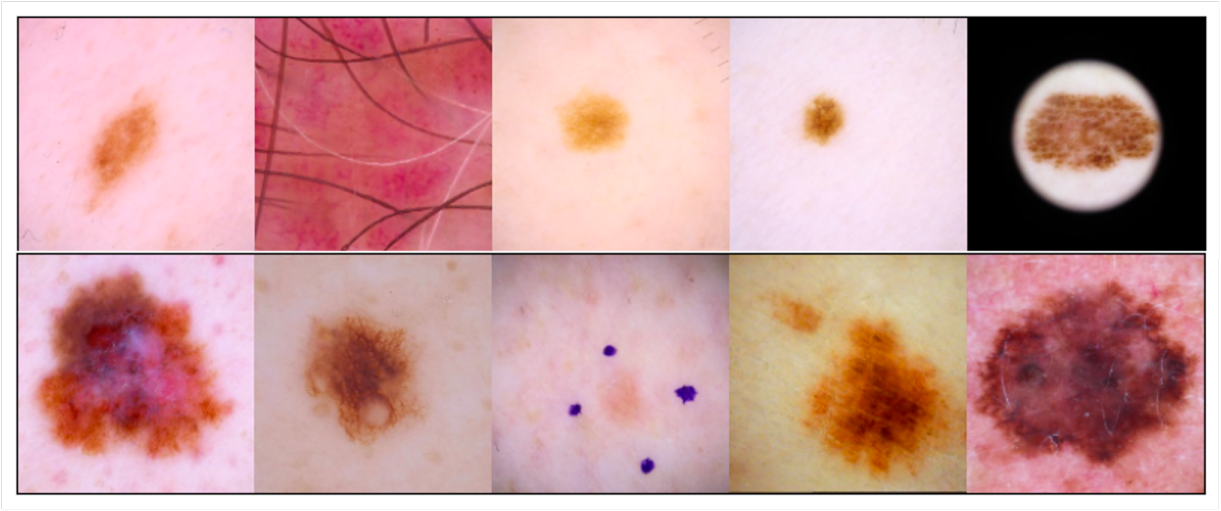}} 
    \hfill
    \subcaptionbox{}{\includegraphics[width=0.49\textwidth]{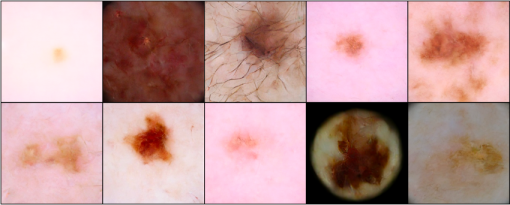}}
    \smallskip

    \subcaptionbox{}{\includegraphics[width=0.49\textwidth]{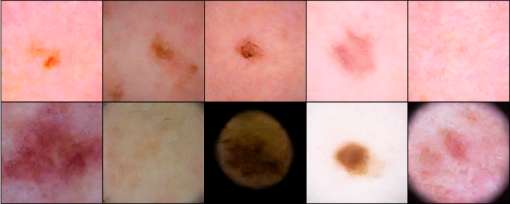}}
    \hfill
    \subcaptionbox{}{\includegraphics[width=0.49\textwidth]{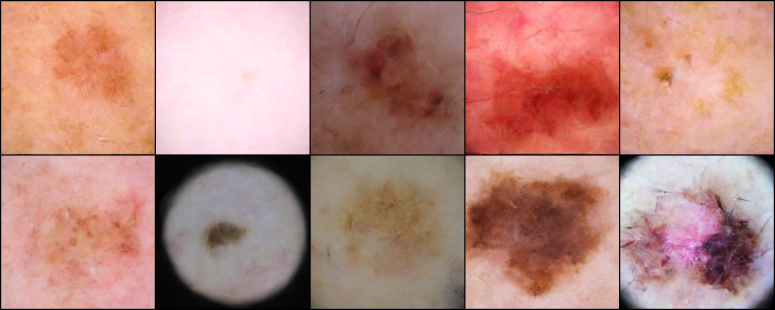}}
    \smallskip

    \subcaptionbox{}{\includegraphics[width=0.49\textwidth]{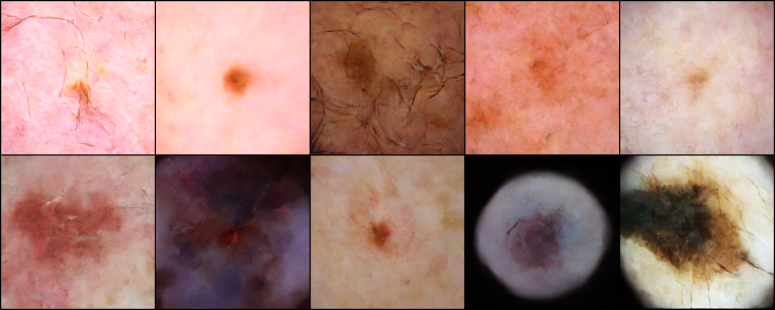}}
    \hfill
    \subcaptionbox{}{\includegraphics[width=0.49\textwidth]{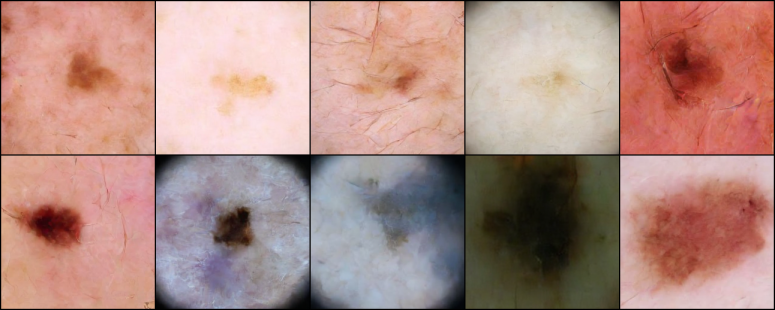}}
    
  \caption{Sample images from (a) ISIC real dataset, (b) generated by DiT without classification model (Setting 1), (c)-(f) generated by the proposed Class-N-Diff (Settings 2-5). For each one, the first row corresponds to benign cases and the second row corresponds to malignant cases.}
  \label{fig:sample_images}

\end{figure*}

\subsection{Inference}
We sample a random noise $z_{t_{max}} \sim \mathcal{N}(0, I)$ from the normal distribution and pass it to the Diffusion Transformer model with the label embedding $y$ to sample $ z_{t-1} \sim p_\theta (z_{t-1} \mid z_t) $. Similar to DiT \cite{peebles2023scalable}, we use $t_{max}$ = 250 sampling steps. After denoising steps from the Diffusion Transformer model, we get $z$, and we use the pre-trained VAE decoder to map it to a realistic dermoscopic image $\hat{x}$.

 \begin{figure*}[t]
    \centering
    \resizebox{\linewidth}{!}{

    \begin{tabular}{ccccc} 
            \includegraphics[width=0.4\linewidth]{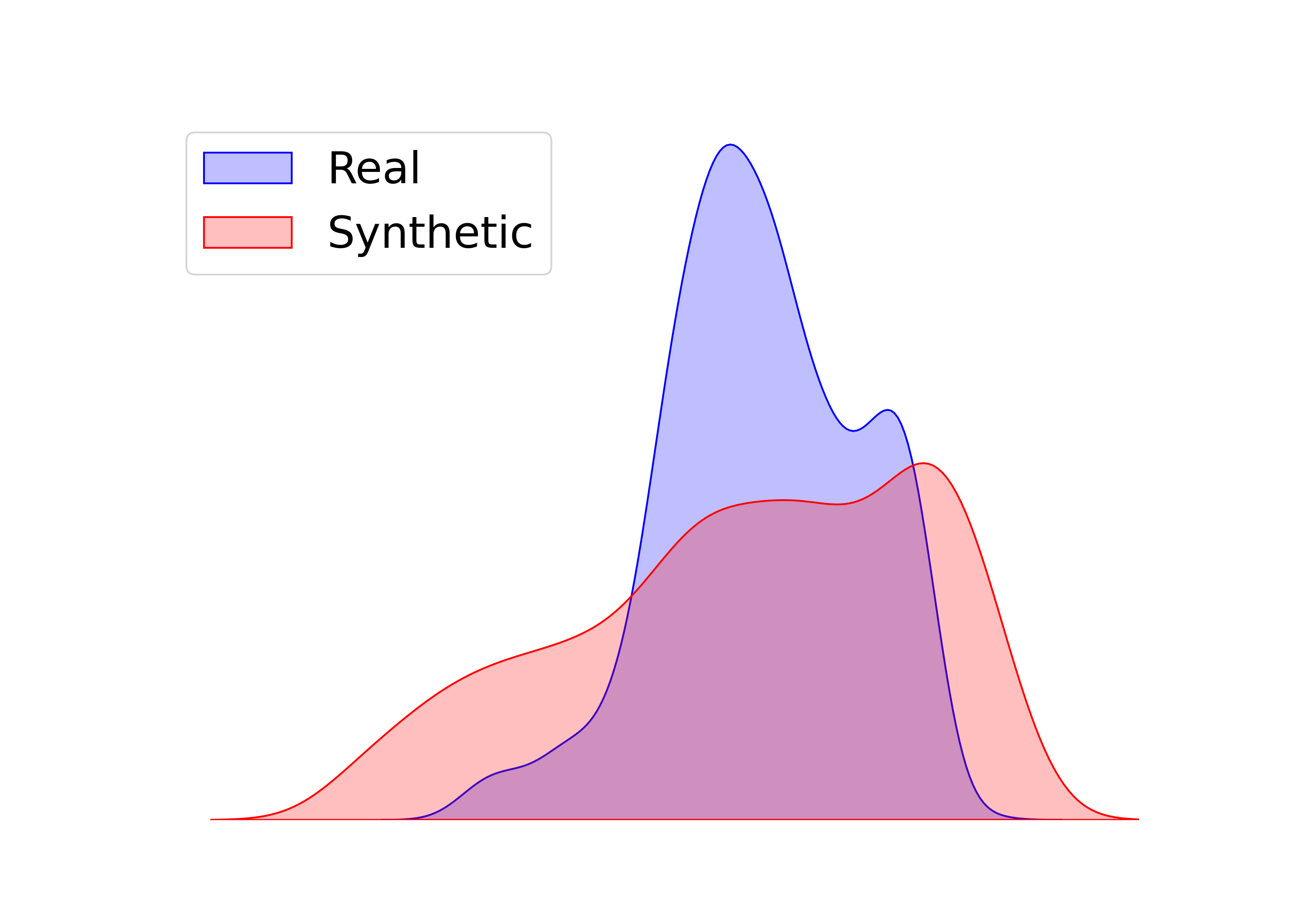} &
            \includegraphics[width=0.4\linewidth]{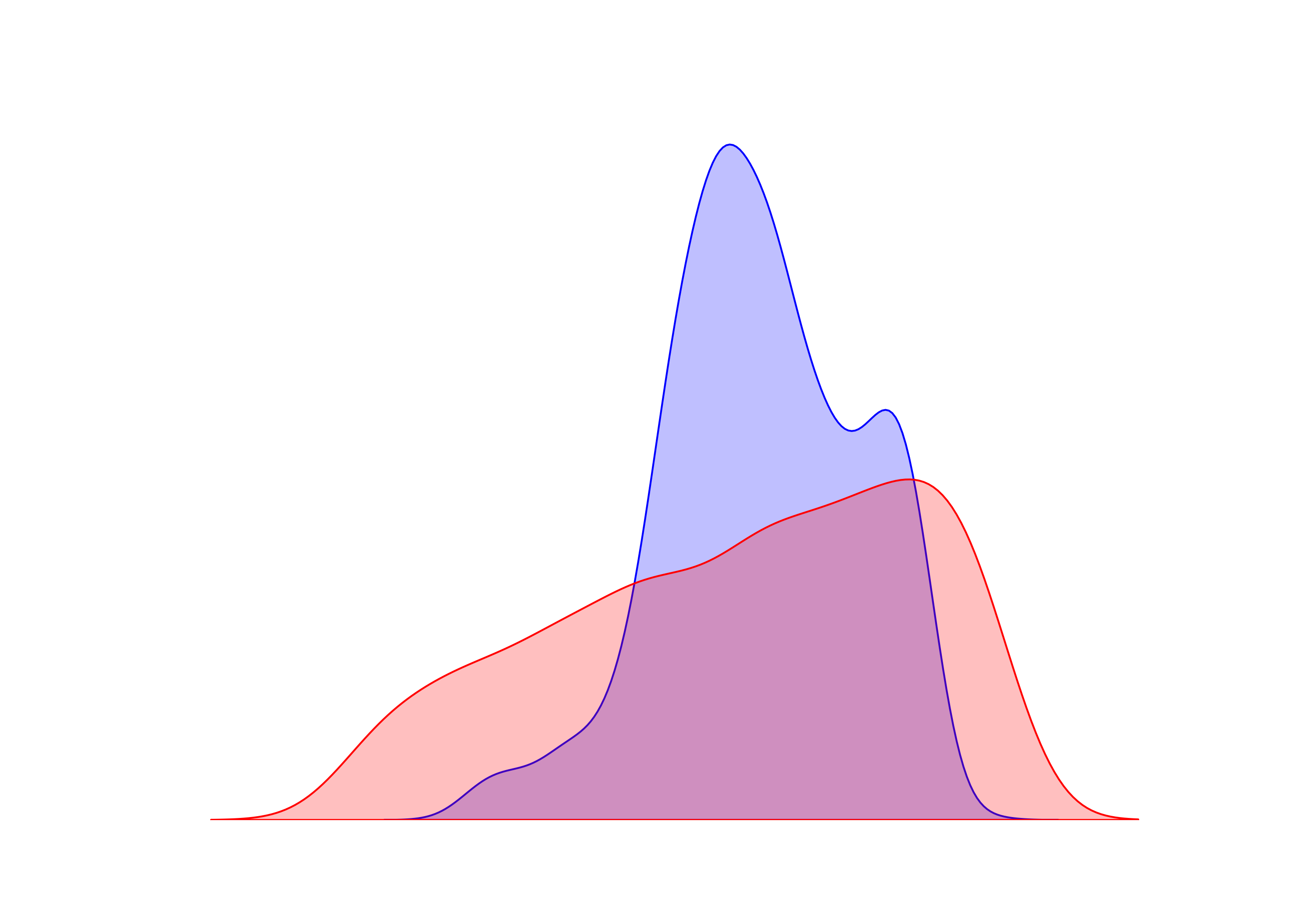} &
            \includegraphics[width=0.4\linewidth]{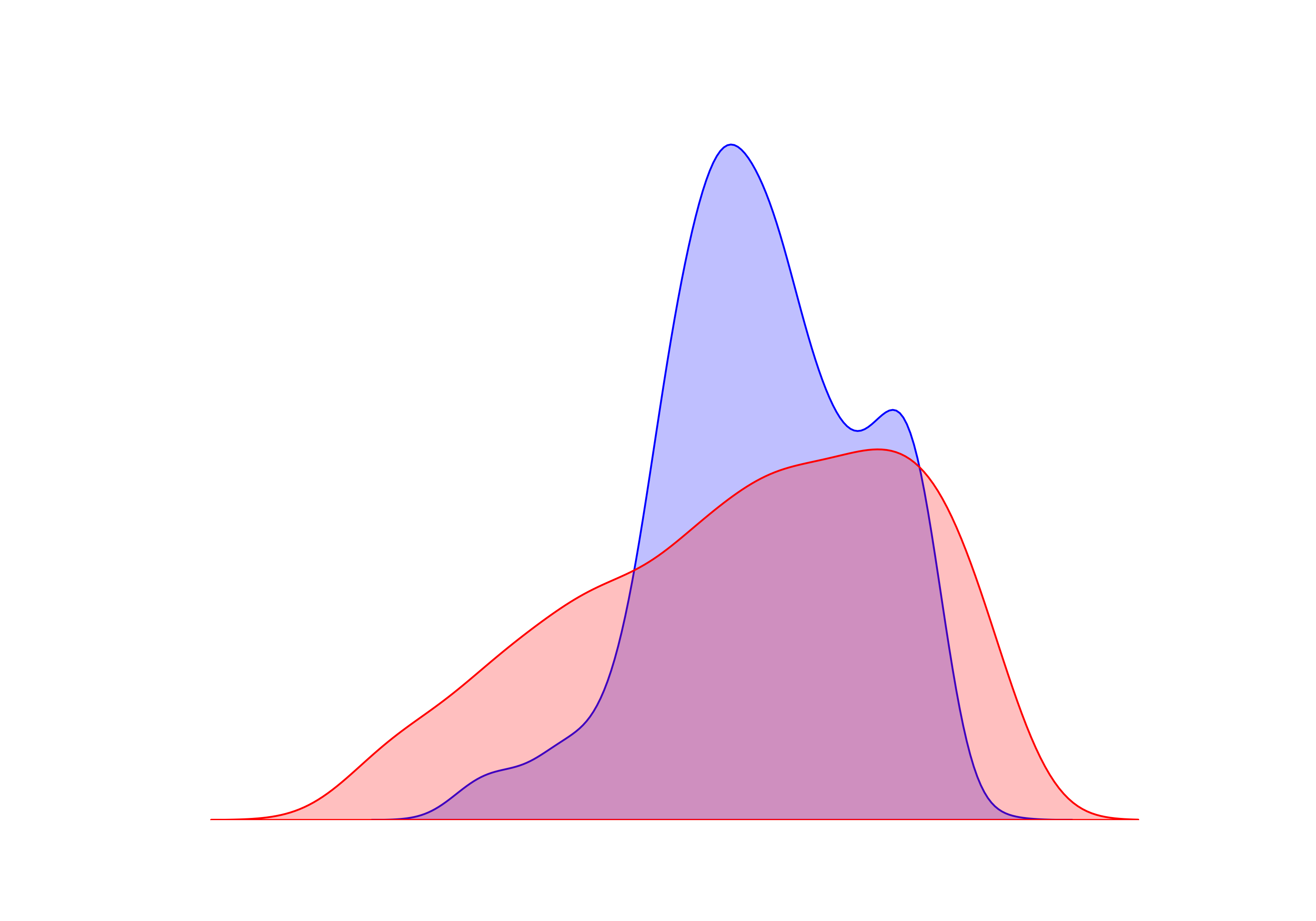} &
            \includegraphics[width=0.4\linewidth]{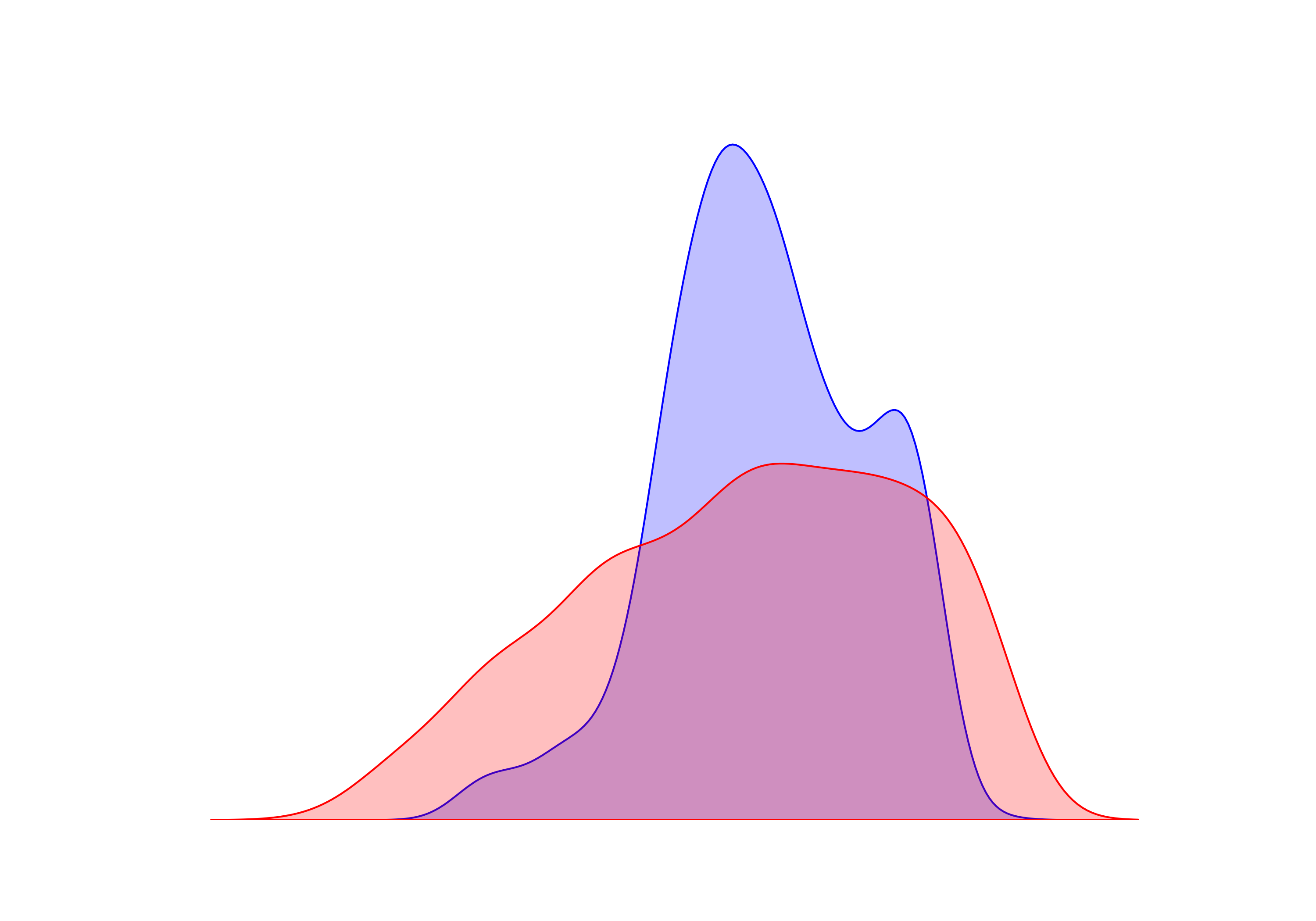}  & 
            \includegraphics[width=0.4\linewidth]{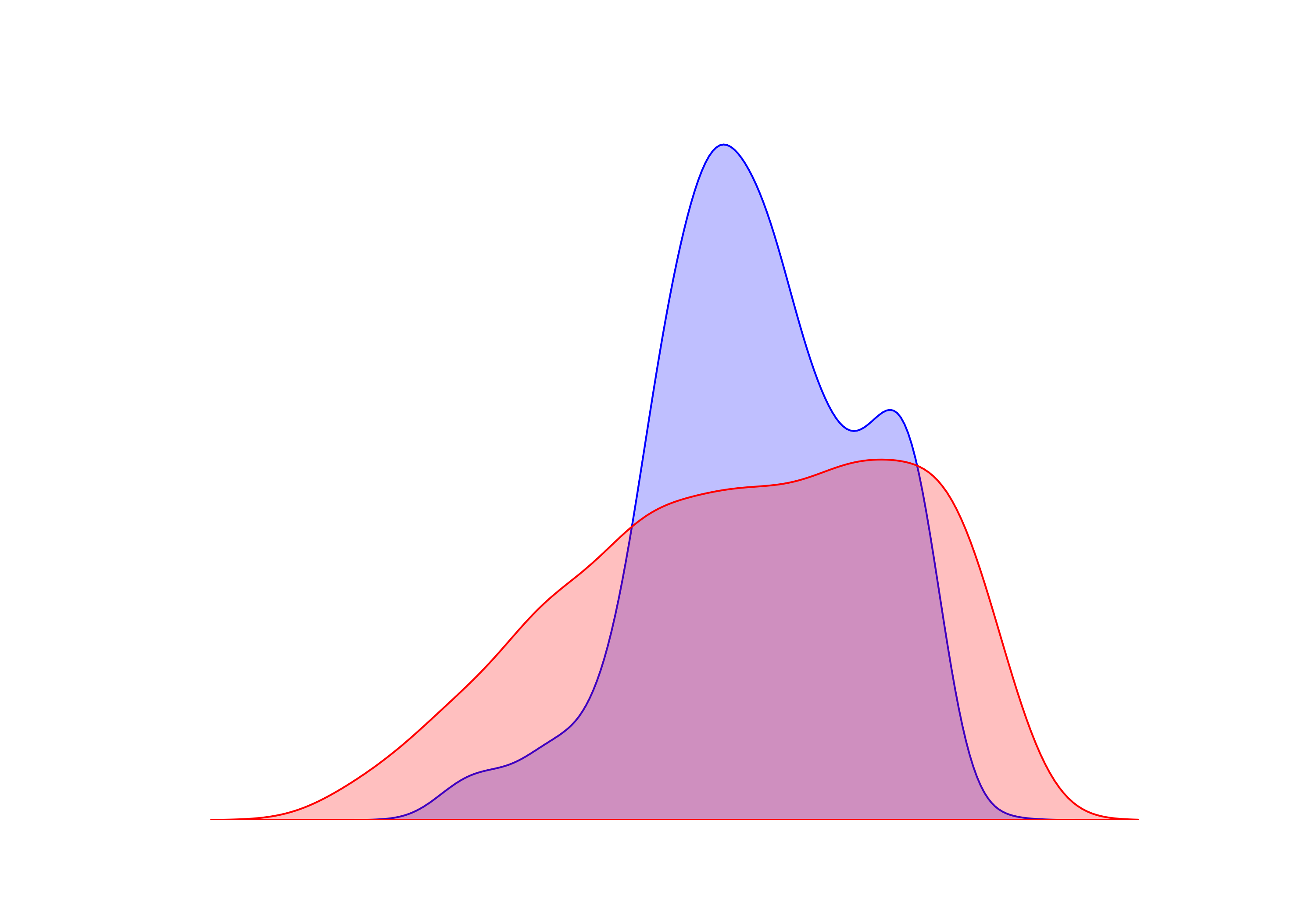} \\
            \large{\textbf{I (a)}} & \large{\textbf{I (b)}} & \large{\textbf{I (c)}} & \textbf{I (d)} & \large{\textbf{I (e)}}\\
        \end{tabular}
    }

    \resizebox{\linewidth}{!}{

    \begin{tabular}{ccccc} 
        \includegraphics[width=0.4\linewidth]{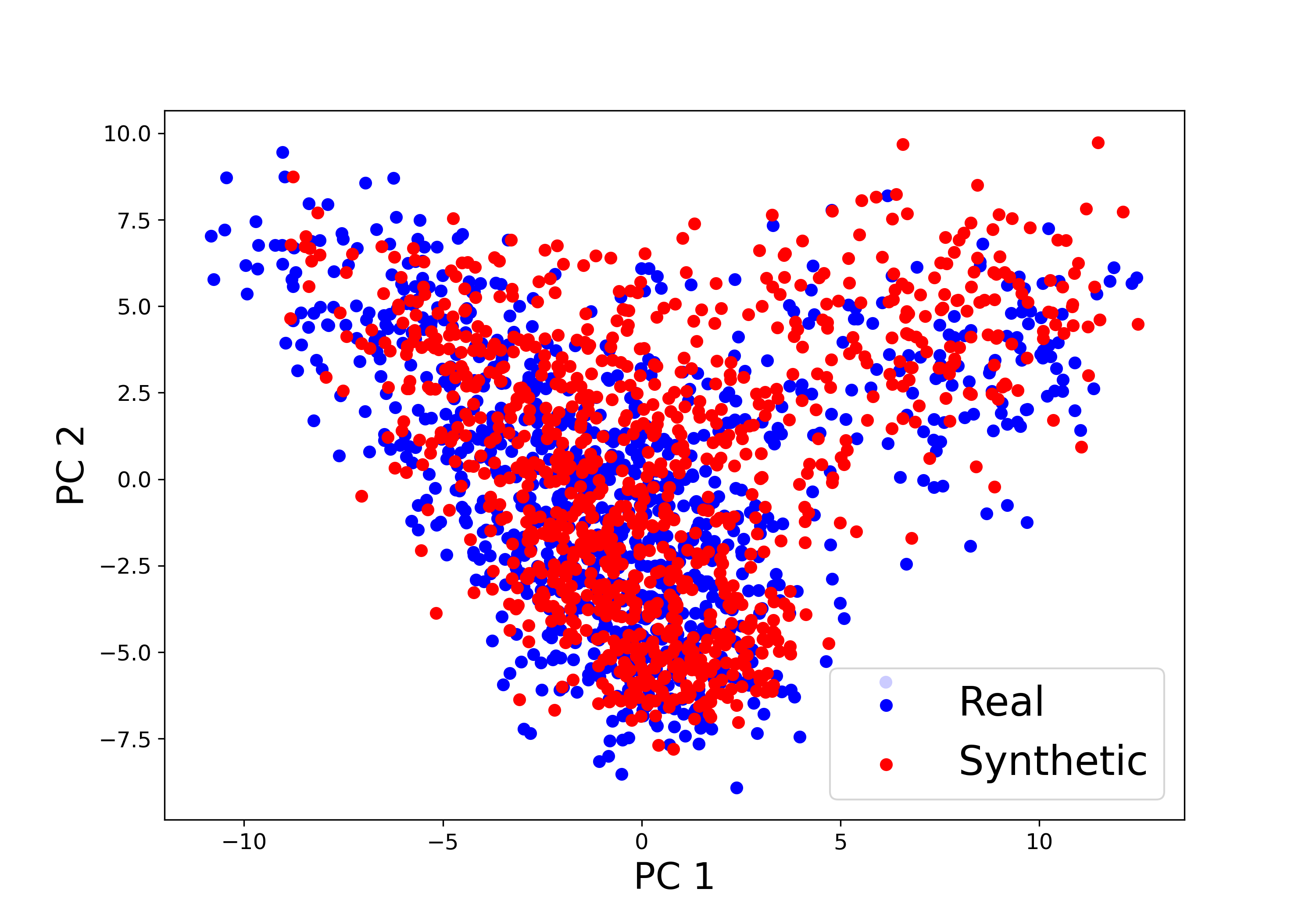} &
        \includegraphics[width=0.4\linewidth]{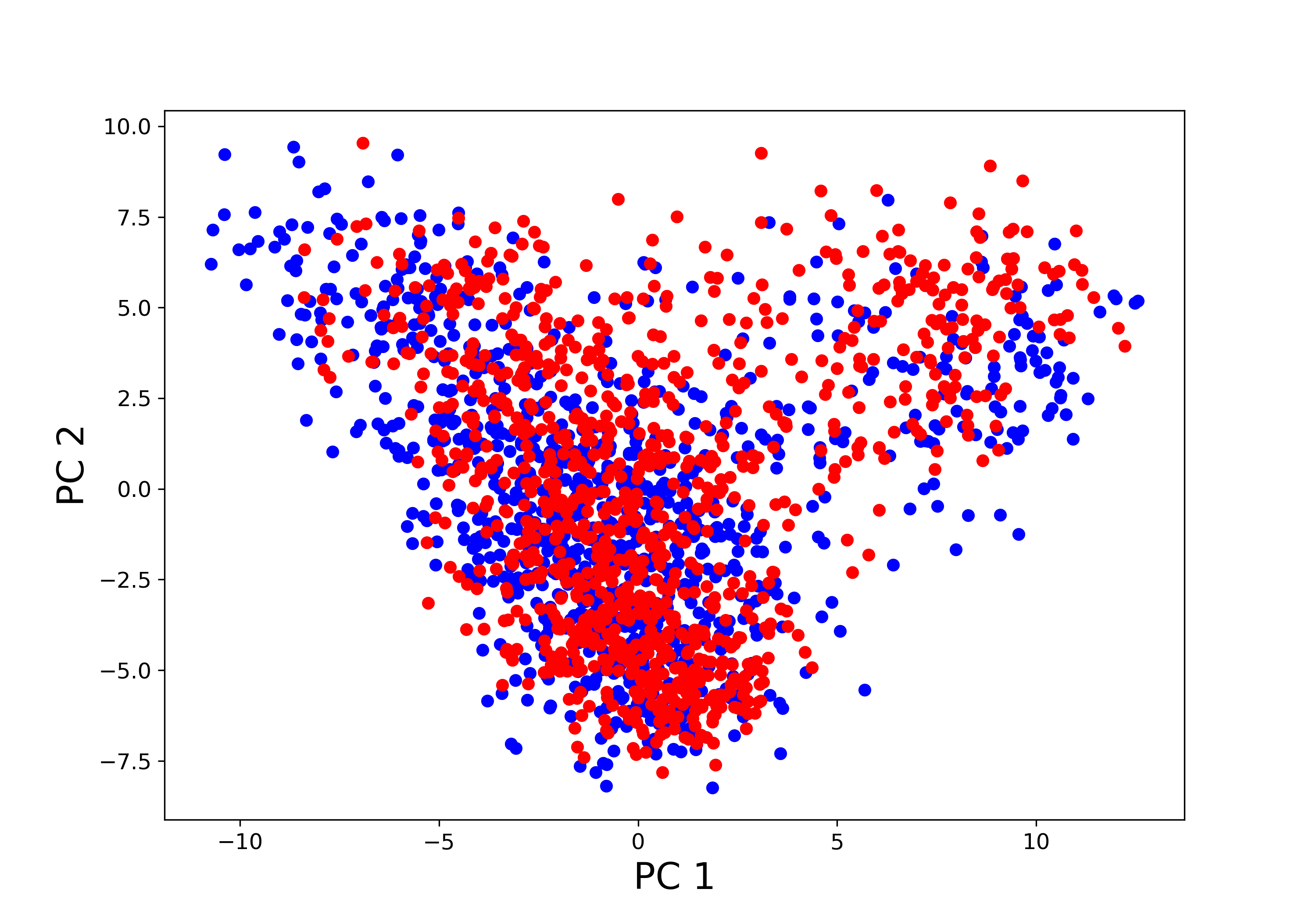} &
        \includegraphics[width=0.4\linewidth]{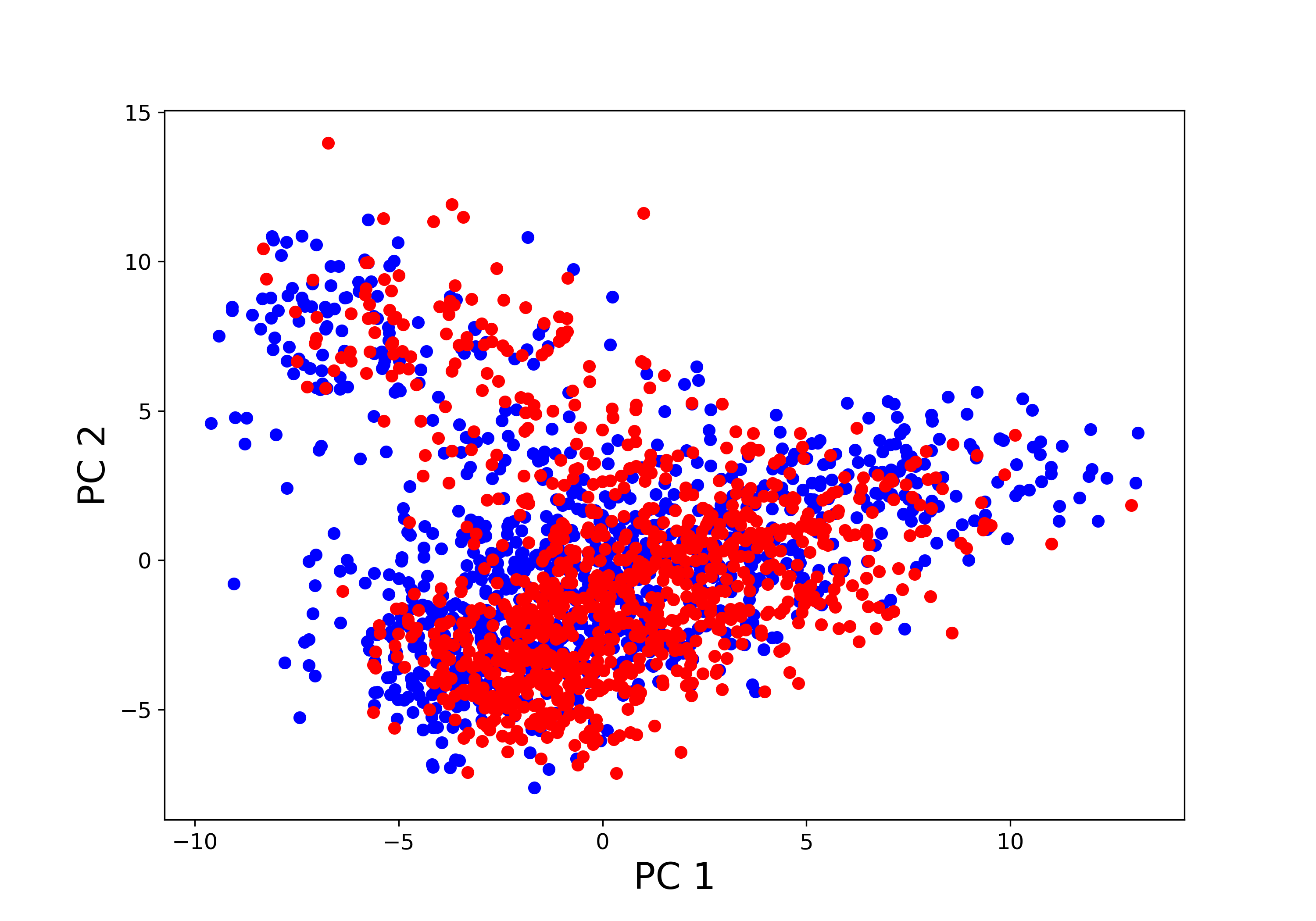} &
        \includegraphics[width=0.4\linewidth]{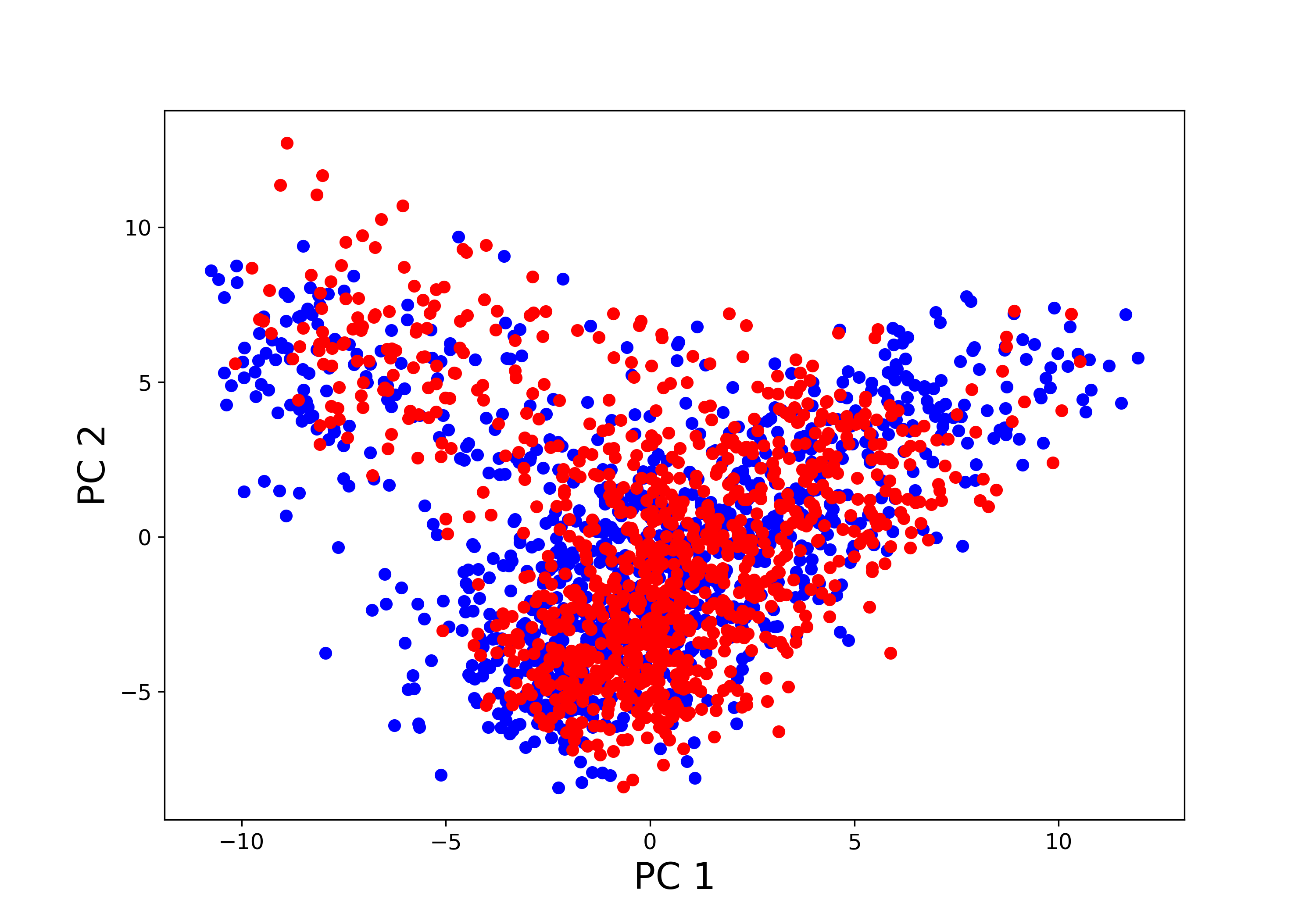}&
        \includegraphics[width=0.4\linewidth]{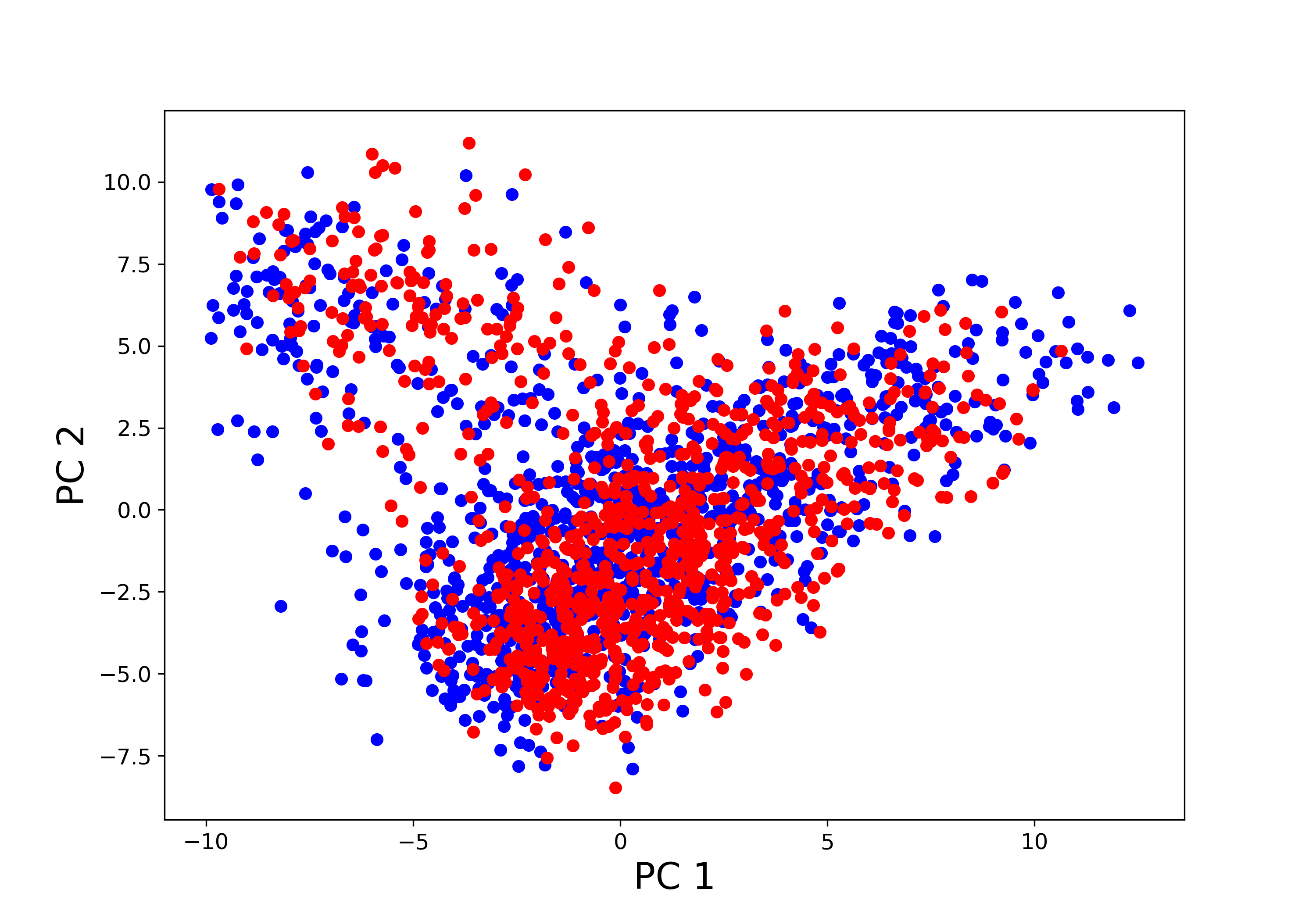}\\
        \large{\textbf{II (a)}}
& \large{\textbf{II (b)}} & \large{\textbf{II (c)}} & \textbf{II (d)} & \large{\textbf{II (e)}}
    \end{tabular}
}
       
\caption{Class-N-Diff Generative performance evaluation: Visualization of the density plots (I) and Principal Components (PCs) (II) to compare the real ISIC data and the synthetic data generated by (a) DiT without classification model (Setting 1), (b)-(e) generated by DiT with classification model (Setting 2-5).}
    \label{fig:all_graphs}
\end{figure*}

\section{Experiments and Results}
\subsection{Implementation Details}
\textbf{Data:} To evaluate the proposed Class-N-Diff model, we used the International Skin Imaging Collaboration (ISIC) datasets [2016-2020]~\cite{codella2018skin, tschandl2018ham10000, combalia2019bcn20000, rotemberg2021patient}. In total, these datasets contain 57960 dermoscopic images and their corresponding gold-standard disease diagnostic metadata. We assessed the downstream classification performance on additional datasets for external validation and fairness evaluations. The Diverse Dermatology Images (DDI) dataset~\cite{daneshjou2022disparities} is a diverse dataset containing a total of 656 dermatology images of three different skin tone categories based on Fitzpatrick skin types (FST)~\cite{fitzpatrick1988validity}. Following~\cite{munia2025dermdiff}, we categorized skin tones into three groups for standardized classification and comprehensive analysis: FST I-II (lighter skin tones) as Type A, FST III-IV as Type B, and FST V-VI (darker skin tones) as Type C. The Fitzpatrick17k dataset~\cite{groh2021fitzpatrick} comprises clinical images of various skin conditions, including annotations of FST skin types. Additional datasets considered for downstream evaluation include Atlas~\cite{kawahara2018seven}, ASAN~\cite{han2018classification}, and MClass~\cite{brinker2019comparing}.

\textbf{Inputs:} All the input dermoscopic images were resized to $256\times256$ resolution and normalized to the range [0,1].

\textbf{Training:} All the models were trained on 57,882 dermoscopic images along with their corresponding class labels from the ISIC datasets. Of them, 52,792 are benign and 5,090 are malignant cases. We implemented our models in Python with the PyTorch library. We trained all models using Dual NVIDIA RTX 4000 GPUs, each equipped with 16 GB of memory.

\textbf{Hyper-parameters:} We trained our generative diffusion model with a mini-batch size of 8 and a learning rate of $1e^{-4}$ for 200k steps. Five different settings were experimented with for the Class-N-Diff model~(Table~\ref{tab:experimental_settings}).

\textbf{Evaluation:} To evaluate the generative performance of our Class-N-Diff model, we calculated Fr\'echet Inception Distance (FID)~\cite{heusel2017gans} and Multi-scale Structural Similarity Index Measure (MS-SSIM)~\cite{wang2003multiscale} scores. Although the FID score is traditionally computed using the ImageNet pre-trained Inception-v3 model, we calculated FID scores based on the ISIC fine-tuned Inception-v3 model~\cite{munia2025dermdiff}. We report the classification performance by calculating accuracy, AUC, and sensitivity scores. 

\begin{table*}[t]
\centering
\caption{Evaluation of the proposed Class-N-Diff in diagnosing skin cancer across three different settings by calculating accuracy, AUC, and Sensitivity scores. }
\label{tab:classification_diffusion}
\resizebox{0.95\linewidth}{!}{%
\begin{tabular}{@{}lc ccc c ccc c ccc@{}} 
\toprule
\multirow{2}{*}{Test Data} & \phantom{a} & 
\multicolumn{3}{c}{Setting 1 (Separate classifier)} & \phantom{a} & \multicolumn{3}{c}{Setting 2 (Class-N-Diff)} &  \phantom{a} & \multicolumn{3}{c}{Setting 3 (Class-N-Diff)} \\ 
\cmidrule{3-5} \cmidrule{7-9} \cmidrule{11-13}
&& Accuracy & AUC & Sensitivity 
&& Accuracy & AUC & Sensitivity 
&& Accuracy & AUC & Sensitivity \\ 
\midrule
DDI\_all  
&& 0.716 & 0.586 & 0.111                
&& 0.732 & 0.590 & 0.088    
&& 0.730 & 0.564 & 0.029 \\

DDI\_A 
&& 0.755 & 0.579 & 0.082                
&& 0.769 & 0.598 & 0.082    
&& 0.755 & 0.522 & 0.000 \\ 
DDI\_B 
&& 0.685 & 0.616 & 0.135                
&& 0.689 & 0.648 & 0.081    
&& 0.693 & 0.563 & 0.027 \\ 
DDI\_C            
&& 0.715 & 0.562 & 0.104               
&& 0.744 & 0.516 & 0.104    
&& 0.749 & 0.608 & 0.062 \\ 
ISIC-2018     
&& 0.882 & 0.769 & 0.070                
&& 0.878 & 0.797 & 0.082    
&& 0.888 & 0.833 & 0.117 \\ 
Fitzpatrick17k     
&& 0.499 & 0.556 & 0.099                
&& 0.509 & 0.567 & 0.086    
&& 0.494 & 0.535 & 0.029 \\ 
AtlasDerm          
&& 0.762 & 0.774 & 0.190                
&& 0.770 & 0.768 & 0.187    
&& 0.787 & 0.799 & 0.179 \\ 
AtlasClinic        
&& 0.748 & 0.659 & 0.040                
&& 0.759 & 0.677 & 0.060    
&& 0.756 & 0.676 & 0.032 \\ 
ASAN               
&& 0.923 & 0.805 & 0.085                
&& 0.897 & 0.724 & 0.085    
&& 0.930 & 0.754 & 0.034 \\ 
MClassDerm            
&& 0.830 & 0.744 & 0.200                
&& 0.820 & 0.744 & 0.150    
&& 0.830 & 0.701 & 0.150 \\ 
MClassClinic            
&& 0.840 & 0.787 & 0.200                
&& 0.830 & 0.829 & 0.250    
&& 0.820 & 0.858 & 0.100  \\ 
\bottomrule
\end{tabular}%
}
\end{table*}

\subsection{Results and Discussion}
\paragraph*{Image Generation Performance}
To evaluate the performance of our proposed Class-N-Diff generative model, we generated a total of 30,000 samples. We randomly picked 5k, 10k, and 20k from the samples and calculated the FID scores with the fine-tuned Inception-v3 model~\cite{munia2025dermdiff}. Table~\ref{tab:fid_scores} reports the FID and MS-SSMIM scores for the baseline DiT model and our proposed DiT with the integrated classification models (Class-N-Diff). As is evident, the Class-N-Diff model has lower FID scores and MS-SSIM scores compared to the original class-conditioned DiT model. Lowest FID is observed when the classification loss weight $\lambda$ is set to 0.2 (Settings 3 and 4). On the other hand, $\lambda$=0.3 in Setting 5 results in the lowest MS-SSIM score. Class-N-Diff consistently performs better than the diffusion-only model across all the settings. The classification loss, combined with the diffusion loss with weighted parameters, enhances the diffusion model to generate more diverse synthetic images. This demonstrates the usefulness of the classification model in the diffusion model. The classification loss pushes features apart in the latent space and forces the reverse diffusion process to respect class boundaries. It directly informs the diffusion network of which features are important for each class. This yields sample images generated by the diffusion model that are both more realistic and diverse, covering all class labels. This is confirmed with the visual comparison of the real ISIC images and generated ones from the five different settings (see Fig.~\ref{fig:sample_images}). A similar trend is observed in the Kernel Density Estimation (KDE) plots and the first two Principal Components (PCs)/features as in Fig.~\ref{fig:all_graphs}.  We randomly sample 1000 images from both real and synthetic images, and plot their data distributions. Although the KDE density plots look almost similar for all settings for training, the PCA plots show how well their data distribution matches the real data distribution.

\paragraph*{Classification Performance}
We train the classification model independently and compare its performance with the model that we trained jointly (Class-N-Diff). Then we test these two models on different in-domain and out-of-distribution test datasets. Table~\ref{tab:classification_diffusion} reports the classification accuracy, AUC, and sensitivity scores. The in-domain test dataset (ISIC-2018) has better accuracy and AUC scores when tested with the classification model jointly trained during the diffusion process. The training of the classification model included real data and also data from the diffusion model, which helped the model to generalize well with diverse dermoscopic images. We also test these classification models on out-of-distribution dataset, DDI, where we report results for each skin tone type: A, B, and C. The classification model trained with the diffusion model (Class-N-Diff) performs better than the original classification model across all three skin tones. For other test sets (Fitzpatrick, Atlas, Asan, and MClass), we also observe a similar pattern (Table~\ref{tab:classification_diffusion}). The diffusion process learning helps the classification model's robustness across diverse dermoscopic data. This classification approach can be expanded into a multi-class framework by incorporating additional demographic attributes such as skin tone and gender. This extension could enhance the versatility of the class-conditioned diffusion model, enabling more diverse and representative dermoscopic image generation.

\section{Conclusions}

We have introduced a classification-induced diffusion framework, Class-N-Diff, that integrates a convolutional classifier with a diffusion transformer to simultaneously perform image synthesis and classification. Our experimental evaluations revealed that joint training consistently lowers FID score relative to the baseline DiT, which confirms that the classification loss effectively guides the denoising process toward higher quality image generation. This integrated approach not only advances generative modeling in medical imaging but also yields a robust classifier trained on both real and synthetic data in an end-to-end fashion. Additional evaluation reveals a marked increase in sample diversity, as evidenced by reduced MS-SSIM scores. Overall, these results demonstrate that our classification-guided diffusion approach is a robust and effective method for generating representative synthetic images while enabling fairer and more accurate skin cancer diagnosis.
\balance
\bibliographystyle{IEEEtran}
\bibliography{refs}

\end{document}